%% file: main.tex
\documentclass[conference]{IEEEtran}
\IEEEoverridecommandlockouts

\usepackage{amsmath,amssymb,amsthm}
\usepackage{graphicx}
\usepackage{epstopdf}
\usepackage{cite}
\usepackage{algorithm}
\usepackage{algpseudocode}
\usepackage{subcaption}
\usepackage[font=footnotesize]{caption}

\usepackage{tikz}
\usepackage{pgfplots}
\usepgfplotslibrary{fillbetween}
\usepgfplotslibrary{groupplots}

\input{latex_resources/mysymbol.sty}
\input{latex_resources/my_sections.tex}

\input{latex_resources/my_beamer_defs.sty}
\input{latex_resources/dsfont.sty}

\theoremstyle{plain}

\newtheorem{claim}{Claim}
\theoremstyle{definition}

\theoremstyle{remark}

\newcommand{\calC}{\mathcal C}
\newcommand{\calP}{\mathcal P}
\newcommand{\calL}{\mathcal L}

\title{Multi-task Supervised Learning via Cross-learning
\thanks{$1$: University of Pennsylvania, United States, e-mails: $\{$jcervino, aribeiro$\}@$seas.upenn.edu. $2$: Universidad de la Rep\'ublica, Uruguay, e-mail: jbazerque$@$fing.edu.uy. $3$: Massachusetts Institute of Technology, e-mail: cfullana$@$mit.edu. This work is supported by NSF-Simons MoDLTheorinet and Uruguay's ANII FSE 1-2019-1-157459.
}
\author{
\IEEEauthorblockN{Juan Cervi\~no$^1$, Juan Andr\'es Bazerque$^2$, Miguel Calvo-Fullana$^3$ and Alejandro Ribeiro$^1$}}
}

\begin{document}

\maketitle

\begin{abstract}
 In this paper we consider a problem known as multi-task learning, consisting of fitting a set of classifier or regression functions intended for solving different tasks.  In our novel formulation, we couple the parameters of these functions, so that they learn in their task specific domains while staying close to each other. This facilitates cross-fertilization in which data collected across different domains help improving the learning performance at each other task. First, we present a simplified case in which the goal is to estimate the means of two Gaussian variables, for the purpose of  gaining some insights on the advantage of  the proposed cross-learning strategy. Then we provide a stochastic projected gradient algorithm to perform cross-learning over a generic loss function. If the number of parameters is large, then the projection step  becomes computationally expensive. To avoid this situation, we derive a primal-dual algorithm that exploits the structure of the dual problem, achieving a formulation whose complexity only depends on the number of tasks. Preliminary numerical experiments for image classification by neural networks trained on a dataset divided in different  domains corroborate that the cross-learned function outperforms both the task-specific and the consensus approaches.
\end{abstract}

\begin{IEEEkeywords}
	Supervised learning, multi-task learning, optimization.
\end{IEEEkeywords}

\section{Introduction}

Supervised learning is one of the traditional problems studied in statistical learning. At its core, it consists of learning  a function mapping inputs to outputs based on provided input-output pairs \cite{hastie2009elements}. In practical situations, it is often the case that in order to learn this function (which we also refer as a \emph{task}), we have limited input-output pairs available (called \emph{training data}). It is also common for different tasks to be related in some sense, and thus, one could attempt to exploit this relationship to improve the individual tasks' performance. The study of this problem, in general is known as \emph{multi-task learning} \cite{caruana1997multitask}. While expected in the limited sample regime, it is also applicable in the asymptotic case, as exemplified by Stein's paradox \cite{stein1956inadmissibility}, which shows that when estimating more than three parameters from Gaussian random variables, a combined estimator exists that has lower Mean Square Error (MSE) than any separate estimator, even if the random variables are independent.

Multi-task learning has been widely used in practice, specially finding success in computer vision \cite{liu2019end,misra2016cross,kendall2018multi}, and language applications \cite{liu2017adversarial,dong2015multi}. Applications aside, approaches to multi-task learning can be grouped in two categories, based on whether they explicitly model or not the relationship between tasks \cite{zhang2017survey}. Out of those assuming a priori knowledge of the tasks, prevalent approaches are based on a commonly shared underlying task representation \cite{ben2003exploiting}. This underlying task representation often takes the form of a sparse, low rank representation across tasks \cite{argyriou2008convex,evgeniou2004regularized}, or a manifold representation \cite{agarwal2010learning}. Some other approaches do not assume previous knowledge, and learn the relationship between tasks directly from data, generally by performing clustering \cite{jacob2009clustered}. In the context of support vector machines, some works have bounded the pair-wise difference between the elements of the classifiers' weights \cite{kato2008multi,kato2009conic}. Some other works focus on finding Pareto optimal solutions between tasks \cite{NEURIPS2018_432aca3a}.

In this paper, we take a constrained approach to the multi-task learning problem. Our formulation is based on the cross-learning framework \cite{cervino2019meta,cervino2020multi}. Originally used for learning policies in a reinforcement learning scenario, its principles can be also applied to the multi-task supervised learning problem. The proposed cross-learning approach consists on bounding the distance between the parameters of the learned functions (in our case, classifiers or regression functions). We show, via an illustrative Gaussian sample mean example, that the use of this centrality measure can guarantee to outperform both the separate and consensus approaches. Furthermore, since the cross-learning method is constrained, it requires a projection at each step which can be computationally expensive (e.g., for neural networks with millions of parameters). To overcome this hurdle, we propose to solve the projection step on the dual domain, which reduces the dimension of the problem to the number of tasks. Finally, we present numerical results in a complex classification task, which show that our cross-learning strategy outperforms the consensus and agnostic estimators.

\section{Problem Formulation}

We address the problem of learning $N$  functions from data coming from different tasks. Let function $f:\ccalX \times \Theta \to \ccalY$ be the map between input space $\ccalX \subset \reals^P$ and output space $\ccalY \subset \reals^Q$ parameterized by $\theta\in\Theta \subset \reals^S$. We seek to minimize the loss function $\ell$ given $N$ datasets containing  pairs $(x_{j},y_{j})\in (\ccalX,\ccalY)$ drawn according to a joint probability $p_i(x,y)$, for  $j=1,\ldots, M_i$, where $M_i$ and $p_i(x,y)$ are the number of samples and generating probability corresponding to each of the  $i=1,\ldots,N$  tasks.  Our objective is to obtain functions $f(x,\theta_i)$ that stay close together in the optimization space via the introduction of a measure of centrality $\epsilon$, and central parameters $\theta_g$. Then, we pose the cross-learning problem
\begin{subequations}
	\begin{align*}
	\tag{PCL}
	\label{eq:originalProblemCrosslearning}
	\{ \theta_i^*\}, \theta_g^*=\underset{\{\theta_i\},\theta_g\in\Theta}{\argmin}  \quad   & \sum_{i=1}^{N}  \mbE_{p_i(x,y)} [\ell \left(y,f(x,\theta_i)\right)] \\
	\text{subject to}
	\quad   & \left\| \theta_i - \theta_g \right\| \leq \epsilon, \quad i=1,\ldots,N.
	\end{align*}
\end{subequations}
The selection of the centrality parameter $\epsilon$ entails a trade-off between learning the specific tasks individually or sharing the data corresponding to all tasks. Indeed, if $\epsilon$ is large enough, the constraint is always inactive and the functions are learned separately, being agnostic to the information provided by the data corresponding to the other tasks. On the other extreme, if $\epsilon=0$, all policies are required to be equal and thus the cross-learning setup reduces to consensus. In practice, the best parameter $\epsilon$ to use is not known, and to estimate it one can resort to cross-validation techniques \cite{hastie2009elements}.

Before presenting an algorithm to solve the cross-learning problem, and to gain insights on the advantages proposed formulation \eqref{eq:originalProblemCrosslearning}, we consider the more tractable problem of estimating the means of two Gaussian variables. This example will allow us to show that by selecting a nonzero $\epsilon$ judiciously, the cross-learning formulation shown in \eqref{eq:originalProblemCrosslearning} can outperform both the agnostic and consensus estimators. 

\subsection{An illustrative case of sample means}\label{sec:Gaussian}

Consider the problem of estimating the means $\mu_x$ and $\mu_y$ of two independent real-valued Gaussian random variables, from their samples $X_j\sim \mathcal N(\mu_x,\sigma^2),\ j=1,\ldots, M$ and $Y_j\sim \mathcal N(\mu_y,\sigma^2),\ j=1,\ldots, M$. If $\mu_x$ and $\mu_y$ are estimated separately, being agnostic of the information provided by the other sample, their maximum likelihood estimators are the sample means $\bar X_M$ and $\bar Y_M$, and the corresponding mean squared errors equal $\sigma^2_M:=\sigma^2/M$.

Now, if we know $|\mu_x-\mu_y|=\epsilon_0$ a priori, then we can use the cross-learning  algorithm to trade off bias for variance and improve the estimation accuracy. Indeed, we will prove
\begin{claim}\label{claim:msebound}
The Mean Squared Error (MSE) of the of the estimates obtained with the cross-learning algorithm for $\epsilon=\epsilon_0$ satisfy $\textrm{mse}_{\text{CL}}(\epsilon_0)\leq (3/4) \sigma^2_M$.
\end{claim}
In words, the agnostic estimator can be outperformed by a factor of $3/4$, at least, regardless of the value of $\epsilon_0$ and $\sigma_M$. This is true even when $M\to\infty$, where the agnostic estimator becomes consistent. Although this result assumes exact knowledge of the ground truth $\epsilon=\epsilon_0$, we will characterize the MSE $\textrm{mse}_{\text{CL}}(\epsilon)$ as a continuous function of $\epsilon$, and demonstrate that   there is a range of values around $\epsilon_0$  for which the cross-learning estimator outperforms the agnostic one. Furthermore, we also argue that cross-learning outperforms consensus, corresponding to $\epsilon=0$. Specifically, 

\begin{claim}\label{claim:msezero}
There exists $\epsilon>0$ such that $\textrm{mse}_{\text{CL}}(\epsilon)<  \textrm{mse}_{\text{CL}}(0)$, with strict inequality for all values of $\epsilon_0$ and $\sigma_M$.
\end{claim}

In proving these claims, we will assume without loss of generality that $\mu_x=\mu_y+\epsilon_0$. Specifying the cross-learning estimator in \eqref{eq:originalProblemCrosslearning} for the problem at hand, we have
\begin{subequations}
	\begin{align}
	(\hat \mu_x^{\text{CL}}, \hat \mu_y^{\text{CL}})=\underset{\bar \mu_x,\bar \mu_y}{\argmin}  \quad   & 
	\sum_{i=1}^M\left[\left(X_i-\bar \mu_x\right)^2+\left(Y_i-\bar \mu_y\right)^2\right] \hspace{-0.5ex} \\
	\text{subject to}
	\quad   & |\bar \mu_x-\bar \mu_y|\leq\epsilon
	\end{align}
\end{subequations}
In this case, the problem admits the following closed-form solution, 
\begin{align}
\hat \mu_x^{\text{CL}}=\begin{cases}
\bar X_M, &\textrm{ if }|\bar X_M-\bar Y_M|<\epsilon\\
(\bar X_M+\bar Y_M)/2+\epsilon/2, & \textrm{ if } \bar X_M>\bar Y_M+\epsilon  \\
(\bar X_M+\bar Y_M)/2-\epsilon/2,  &\textrm{ if } \bar X_M<\bar Y_M-\epsilon  
\end{cases}\label{eq:muxhat}
\end{align}
where the cases correspond to the constraint being activated or not. In order to compute the mean squared error, it is convenient to put $\hat \mu_x^{\text{CL}}$ in terms of the variables $\bar Z_M:=(\bar X_M+\bar Y_M)/2$ and $\bar W_M:=(\bar X_M-\bar Y_M)/2$. Under these definitions, \eqref{eq:muxhat} reduces to $\hat \mu_x^{\text{CL}}=\bar Z_M+T(\bar W_M)$, where 
\begin{align}
T(\bar W_M)&:=\bar W_M \mathds 1[|\bar W_M|<\epsilon/2] \nonumber\\
&+(\epsilon/2) \mathds 1[\bar W_M>\epsilon/2]-(\epsilon/2) \mathds 1[\bar W_M<-\epsilon/2].
\end{align}
Since $\bar X_M$ and $\bar Y_M$ are independent, with the same variance, then $\bar Z_M$ and $\bar W_M$ also are, thus
\begin{align}
        \textrm{mse}_{\text{CL}}(\epsilon)&=\mbE[(\hat \mu_x^{\text{CL}}-\mu_x)^2]\\&
        =\mbE[(\bar Z_M-\mu_z)^2]+\mbE[(\bar T(W_M)-\mu_w )^2]\\
        &=\sigma_M^2/2+ E[(\bar W_M-\epsilon_0/2)^2 \mathds 1[|\bar W_M|<\epsilon/2] \nonumber \\
        &+\mbE[(1/4) (\epsilon-\epsilon_0)^2\mathds 1[\bar W_M>\epsilon/2]] \nonumber \\
        &+\mbE[(1/4) (\epsilon+\epsilon_0)^2\mathds 1[\bar W_M<-\epsilon/2]].
\end{align}

With $\bar Z_M$ and $\bar W_M$ being Gaussian, the expected values can be found in terms of the error function $\textrm{erf}(x):=(2/\sqrt{\pi})\int_{0}^xe^{-t^2}dt$,
\begin{align}
        \textrm{mse}_{\text{CL}}(\epsilon)&=\frac{\sigma_M^2}{2}\left(1+  \frac{\alpha e^{-\alpha^2}-\beta e^{-\beta^2}}{\sqrt{\pi}}+\frac{e(\beta)-\textrm{erf}(\alpha)}{2}\right)\nonumber \\
       & + \frac{ \sigma_M^2}{2}\left(\alpha^2(1+\textrm{erf}(\alpha))+\beta^2(1-\textrm{erf}(\beta))\right),\label{eq:mseCLerr}
\end{align}
with $\alpha:=-(\epsilon+\epsilon_0)/2\sigma_M$ and $\beta:=(\epsilon-\epsilon_0)/2\sigma_M$.

\begin{figure}[t]
	\centering
	\input{fig/fig_mse.tex}
	\caption{Mean squared error with respect to the centrality parameter $\epsilon$ of the cross-learning problem for one variable. The illustrative case for the estimation of the means of two Gaussian variables is shown. The two variables have same variance $\sigma=1$ and the distance between their means is given by $\epsilon_0=2$. }
	\label{fig:MSE_gaussians}
\end{figure}
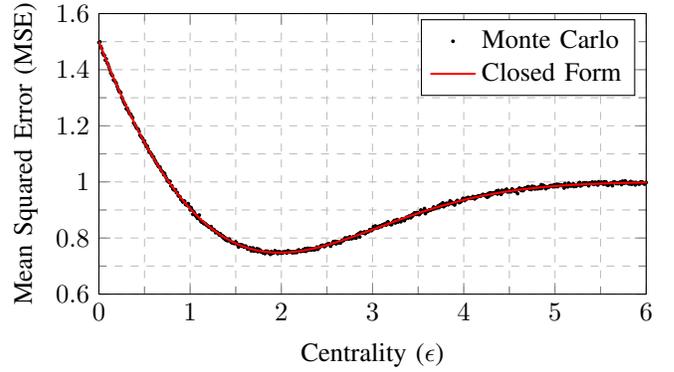

Figure \ref{fig:MSE_gaussians} shows the mean squared error in  \eqref{eq:mseCLerr} depicted in red as a function of $\epsilon$ for a particular choice $\epsilon_0=2$ and $\sigma=1$. Starting from the consensus case at $\epsilon=0$, the error reduces to a minimum value by  $\epsilon_0$, being $3/4$ lower than  the limiting error $\lim_{\epsilon\to\infty}\textrm{mse}_{\text{CL}}(\epsilon)= \sigma_M$. This limit results from the constraint being activated with probability zero, which yields $\hat\mu_x^{\text{CL}}=\bar X_M$, as $\epsilon$ grows unbounded. The red curve in is double checked by comparing it to Monte Carlo estimates of the error, over $100{,}000$ realizations of $\hat \mu_x^{\text{CL}}$,  depicted by black dots in Figure \ref{fig:MSE_gaussians}.
 
To see that Claim \ref{claim:msebound} is true regardless the values of $\sigma_M$ and $\epsilon_0$, put $\epsilon=\epsilon_0$ in \eqref{eq:mseCLerr}, so that $\beta=0$ and  $\alpha=-  \epsilon_0/\sigma_M$, thus
\begin{align}
        \textrm{mse}_{\text{CL}}(\epsilon_0)=    & \frac{\sigma_M^2}{2}\left(1+  \frac{\alpha e^{-\alpha^2}}{\sqrt{\pi}}-\frac{\textrm{erf}(\alpha)}{2}+\alpha^2(1+\textrm{erf}(\alpha))\right).\nonumber
\end{align}

It remains to prove that the term inside brackets in the previous expression does not exceed $3/2$.  Indeed, the function
 $$f(x)=1+\frac{x e^{-x^2}}{\sqrt{\pi}}-\frac{\textrm{erf}(x)}{2}+x^2(1+\textrm{erf}(x))$$ attains $f(-\infty)=3/2$, $f(0)=1$, and $df/dx=2x(1+\textrm{erf}(x))<0$ in the interval $(-\infty,0]$, which  yields $f(x)\in(1,3/2), \forall\ x\in(-\infty,0)$, holding in particular for $x=-  \epsilon_0/\sigma_M=\alpha$, as desired.

 To see Claim \ref{claim:msezero} is also true, we compute the derivative of      \eqref{eq:mseCLerr} with respect to $\epsilon$ and show that it is negative at $\epsilon=0$,
\begin{align}
       D(\epsilon):= \frac{d}{d\epsilon}\textrm{mse}_{\text{CL}}(\epsilon)=    & \sigma_M\left(\beta(1-\textrm{erf}(\beta))-\alpha(1+\textrm{erf}(\alpha))\right).\nonumber
\end{align}

For $\epsilon=0$ $\alpha=\beta=-\epsilon_0/2\sigma_M<0$, hence $D(0)=(\epsilon_0/\sigma_M )\textrm{erf}(-\epsilon_0/2\sigma_M)<0$, which is strictly negative as desired. 

\begin{algorithm}[t]
	\caption{Cross-learning algorithm}
	\label{alg:CLA}
	\begin{algorithmic}[1]
		\State   Initialize $\theta_g=0,$ $\theta^0_i=\theta^0,\ i=1,\ldots,N$
		\Repeat \ for   $k=0,1,\ldots$
		\For {$i=1,\ldots,N$}
		\State  Obtain $\hat \nabla_{\theta_i}\ell(y,f(x,\theta_i^k))$.
		\EndFor
		\State   $(\{\theta_i^{k+1}\},\theta_g^{k+1})=\calP_{\calC}\left[\{ \theta^k_i+\eta^k \hat \nabla_{\theta_i}\ell(y,f(x,\theta_i^k)), \theta_g^k\right]$ 
\Until convergence 
\end{algorithmic}
\end{algorithm}

\section{Algorithm Construction}\label{sec:algorithm}
In order to solve the cross-learning problem \eqref{eq:originalProblemCrosslearning} in a general setup, we can resort to a stochastic projected gradient descent scheme. To take gradient steps on the objective function, we can take derivatives of the loss function with respect to the parameters,
\begin{align}\label{eq:gradient_step}
	\bar \theta_i^{k}=\theta_i^{k}-\eta^k \hat \nabla_{\theta_i} \ell(y,f(x,\theta_i^k)),
\end{align}
where $\hat \nabla_{\theta_i} \ell(y,f(x,\theta_i^k))$ is a stochastic version of the gradient associated to the loss function at a data point, and $\eta^k$ possibly constant, is a learning step.  After each gradient step is taken, the restriction may not be satisfied and thus a projection must be enforced. We define the cross-learning projection $\calP_{\calC}$ as,
\begin{subequations}
	\begin{align}
	\hspace{-0.4cm}\calP_{\calC}\left[\{\bar \theta_i\}, \bar \theta_g\right] =\underset{\{\theta_i\},\theta_g}{\argmin}     & 
	\sum_{i=1}^{N} \left\| \theta_i - \bar\theta_i \right\|^2 +\left\|  \theta_g - \bar \theta_g \right\|^2 \\
	\text{s.t.}
	\quad   & \left\| \theta_i - \theta_g \right\|^2 \leq \epsilon^2, i=1,\ldots,N
	\end{align}
	\label{eq:projectionProblemC}	
\end{subequations}
By combining the gradient step given in equation \eqref{eq:gradient_step} with the projection \eqref{eq:projectionProblemC}, we obtain the cross-learning algorithm illustrated in Algorithm \ref{alg:CLA}. As a projected gradient descent form, it can be shown to converge to the optimal value of the cross-learning problem \eqref{eq:originalProblemCrosslearning} in the case of a convex problems \cite{bertsekas1999nonlinear}. In general, the cross-learning problem \eqref{eq:originalProblemCrosslearning} may not be convex due to both the loss function $\ell$ and the parameterization function $f(x,\theta)$ used. However, recent results have shown that even in those cases, problems akin to \eqref{eq:originalProblemCrosslearning} have tractable duality gaps \cite{chamon2020empirical}, motivating primal-dual approaches. Nonetheless, for the specific case of the cross-learning Algorithm \ref{alg:CLA}, previous results have shown its converge in high probability to a neighborhood of a first-order stationary point of problem \eqref{eq:originalProblemCrosslearning} in the context of reinforcement learning \cite{cervino2020multi}. 

\begin{algorithm}[t]
	\caption{Cross-learning projection}
	\label{alg:CLProj}
	\begin{algorithmic}[1]
		\State   Initialize $\mu_i=\mu_i^0$ and prescribe error $\delta>0$
		\Repeat \ for   $k=0,1,\ldots$
		\For {$i=1,\dots,N$}
		\State $\mu_i^{k+1}=[\mu_i^{k}+\alpha^k \partial_{\mu_i}\calL(\theta_i,\theta_g,\mu^k_i)]_+$ with eq. \eqref{eq:dualProj}
		\EndFor
		\Until  $|\langle \hat \partial_{{\mu}}\calL(\theta_i,\theta_g,\mu^k_i), {\mu}^k \rangle|\leq \delta$
	\end{algorithmic}
\end{algorithm}

\subsection{Projection in the dual domain}
On a neural network, the function parameterization vectors $\theta_i$ may have millions of parameters rendering projection \eqref{eq:projectionProblemC} challenging in practice. However, we can exploit the fact that only one constraint is added per function, resorting to a dual domain algorithm which has only one variable per function. With Lagrange multipliers $\mu_i \geq 0 $, we can write the Lagrangian $\calL(\theta_i, \theta_g,\mu_i)$ of problem \eqref{eq:originalProblemCrosslearning} as,
\begin{align}
	\calL(\theta_i, \theta_g,\mu_i) &=	\sum_{i=1}^{N} \left\| \theta_i - \bar \theta_i \right\|^2 +\left\| \theta_g - \bar \theta_g \right\|^2   \nonumber\\
	&+\sum_{i=1}^{N} \mu_i ( \left\| \theta_i - \theta_g \right\|^2 - \epsilon^2). 
\end{align}
Upon defining $\lambda_i=\frac{\mu_i}{1+\mu_i}$, $z=[1,\lambda_1,\dots,\lambda_N]$ and $a=|z|_1$, the primal minimizers of the Lagrangian given $\mu$ are
\begin{align}\label{eq:primal_solution_g}
	\theta_g&=\frac{1}{a}(\bar{\theta_g}+\sum_{i=1}^{N}\lambda_i \bar{\theta}_i),\\\label{eq:primal_solution_i}
	\theta_i&=(1-\lambda_i)\bar{\theta}_i+\lambda_i \theta_g.	
\end{align}
Hence, we can obtain the \textit{subgradient} of the Lagrangian with respect to the Lagrangian multipliers $\mu_i$, by substituting \eqref{eq:primal_solution_g} and \eqref{eq:primal_solution_i},
\begin{align}
	\partial_{\mu_i}\calL(\theta_i,& \theta_g,\mu_i) 	=\|\theta_i-\theta_g\|^2-\epsilon^2\label{eq:dualProj} \\
	&=(1-\lambda_i)^2 \left\|\bar{\theta}_i-\frac{1}{a}(\bar{\theta_g}+\textstyle\sum_{i=1}^N \lambda_i\bar{\theta}_i)\right\|^2-\epsilon^2.\nonumber
\end{align}
By expanding the norm in  \eqref{eq:dualProj}, the gradient of the Lagrangian depends on the parameters $\bar \theta_i$ through their inner products $\bar \theta_i^T\bar \theta_{i'}$, and these products can be computed once when initializing the primal-dual algorithm. Other than that, Algorithm \ref{alg:CLProj} operates on the reduced dimension of the dual variables.    

The subgradient \eqref{eq:dualProj} is then embedded in the projection described in Algorithm \ref{alg:CLProj}. As $\mu_i$ must be nonnegative \cite{boyd2009convex}, we project them to the nonnegative orthant at every step. The stopping condition of Algorithm \ref{alg:CLProj} is the duality gap of the projection \eqref{eq:projectionProblemC}, which can be made arbitrarily small. Note that the cross-learning projection \eqref{eq:projectionProblemC} is a convex problem, by selecting a non-summable and square summable step-size \cite{bertsekas1999nonlinear}, once the algorithm is halted, the optimal parameters $\{\theta_i\},\theta_g$ of can be recovered using the primal solutions \eqref{eq:primal_solution_g}$-$\eqref{eq:primal_solution_i}. 

\section{Numerical Results}

\input{fig/fig_dataset.tex}

In this section we test our cross-learning framework on a classification problem with real data. Our goal is to classify images belonging to $P$ different categories, and the problem  is divided in $N$ tasks corresponding to images belonging to $N$  different domains. Specifically, we use the Office-Home dataset \cite{venkateswara2017Deep}. It consists of $N=4$ different domains; Art: an artistic representation of the object, Clipart: a clip art reproduction, Product: an image of a product for sale, and Real World: pictures of the object captured with a camera. The overall dataset contains $15{,}500$ RGB images divided in $P=65$ categories, with five examples  given in Figure \ref{fig:Dataset}, including Alarm, Bike, Glasses, Pen, and Speaker. Notice that within each category there are images belonging to each of the domains.  The minimum number of images per domain and category is $15$ and the image size varies from the smallest image size of $18 \times 18$ to the largest being $6500 \times 4900$ pixels. We preprocessed the images by normalizing them and fitting their size to $224 \times 224$ pixels.


We use our cross-learning strategy for classification with the intuition that seeing an image of one domain could help identifying an image of the same category in other domain. For instance, there is a large amount of speakers pictured as products, which could help identify Clipart versions of the speakers which are available in much smaller numbers. We also want to check how the cross-learning classifier compares to  using a single classifier that merges the dataset disregarding the different domains.   
    
We use neural networks as the classifiers $f(x,\theta_i)$, $i=1,\ldots,N$ for cross-learning, with the architecture being based on AlexNet \cite{krizhevsky2012imagenet} with a reduction on the size of the last fully connected layers to $256$ neurons per layer, corresponding to $\theta \in \reals^S$, with $S=4{,}911{,}745$. In this case, it is crucial to make use of the cross-learning projection \eqref{eq:projectionProblemC} in the dual domain, thus reducing the dimensionality from $S$ to $N=4$ variables. Furthermore, we split the dataset in two parts, using  $4/5$ of the images  for training and $1/5$ for testing. We train each  neural network in a domain according to Algorithm \ref{alg:CLA}. For the gradient step, we use a stepsize $\eta=0.001$ and we take one image per step.  As it is standard for image classification, we use the cross-entropy loss \cite{hastie2009elements}. We train the $N=4$ neural networks according to the cross-learning algorithm for different values of the centrality measure $\epsilon$. We also consider the case of consensus ($\epsilon=0$) which is equivalent to merging the images from all domains and training a single neural network. Additionally, we train the neural networks separately one in each domain, which corresponds to $\epsilon=\infty$.

\begin{figure}[t]
\centering
\input{fig/fig_epsilons_mean.tex}
\caption{Accuracy in the test set of a classifier trained with the cross-learning algorithm for different values of the parameter $\epsilon$. Mean values and one standard deviation band are plotted, corresponding to the accuracy over all the $65$ categories and $4$ different domains, obtained after after $30$ epochs. The best overall accuracy is achieved at $\epsilon=10^{-6}$. The consensus case corresponds to $\epsilon=0$ and the agnostic case corresponds to $\epsilon=\infty$ (not plotted, achieving a lower accuracy of $24.44\%$).}
\label{fig:epsilons_mean}
\end{figure}

\begin{figure}[t]
	 \captionsetup[subfigure]{belowskip=4pt, aboveskip=-8pt}
    \centering
	\begin{subfigure}[t]{\columnwidth}
		\centering
		\input{fig/fig_bars_agnostic.tex}
		\caption{\footnotesize{Percentage of improvement in the accuracy of the classifier between the task-specific agnostic classifier ($\epsilon=\infty$) and the best cross-learning classifier. }}
		\label{fig:bar_agnostic}
	\end{subfigure}
	\begin{subfigure}[t]{\columnwidth}
		\centering
		\input{fig/fig_bars_consensus.tex}
		\caption{\footnotesize{Percentage of improvement in the accuracy of the classifier between the task-specific consensus classifier ($\epsilon=0$) and the best cross-learning classifier.}}
		\label{fig:bar_consensus}
	\end{subfigure}
	\caption{Percentage of improvement in the accuracy of the best cross-learning classifier with respect to the agnostic, and consensus classifiers.}
	\label{fig:bars}
\end{figure}

In Figure \ref{fig:epsilons_mean}, we present the result of these experiments using the classification accuracy of the trained classifier as figure of merit. We corroborate the intuitive idea drawn by looking at the images (cf. Figure \ref{fig:Dataset}). Namely, that domains are correlated and adding samples from different domains improves the overall performance of the agnostic policy, only trained on its own specific samples. This is exemplified by consensus ($\epsilon=0$) outperforming the agnostic training ($\epsilon=\infty$), with an accuracy of $33.97\%$ against $24.44\%$. More importantly, the appropriate choice of the $\epsilon$ parameter in the cross-learning method outperforms both consensus and agnostic approaches. In particular, Figure \ref{fig:epsilons_mean} shows that the maximum performance is achieved at $\epsilon=10^{-6}$, indicating that keeping the parameters of the neural networks close, yet not merging them, induces an improvement in the overall accuracy. In this sense, these experimental results with real data recover our theoretical findings about the behavior of the cross-learning estimator in the Gaussian example of Section \ref{sec:Gaussian}.
 
The advantage of cross-learning can be better seen in Figure \ref{fig:bars}, as in all $4$ domains there exists a value of $\epsilon$ that outperforms both the consensus and agnostic counterparts across all of them. In particular, for the domain that has the minimum number of samples (Art, with $2{,}427$), the accuracy almost triples with cross-learning when compared with the agnostic classifier. This should not come as a surprise, as due to its limited number of samples, it admits a larger margin of improvement by including samples from additional domains. Likewise, considerable improvements are still found when compared with the consensus classifier.

\section{Conclusion}

In this paper, we introduced a cross-learning framework for multi-task supervised learning. The goal is to learn a set of functions, either for classification or regression, corresponding to different tasks. The proposed strategy entails the fitting of the parameters of each function to their task-specific data, while keeping the parameters of all functions close to each other. We derived intuition from an illustrative case with Gaussian distributions, corroborating the advantage of our formulation compared both to the consensus or the separate approach. For generic set tasks, we have derived a stochastic projected gradient algorithm, together with a dual implementation of the projection for reducing the problem dimension to the number of tasks. Numerical experiments on a dataset with different domains validate that the cross-learning approach is capable of outperforming both the domain specific and the consensus classifiers. 

\vspace{-0.15cm}
\bibliographystyle{IEEEtran}
\bibliography{bib}

\end{document}

%% file: latex_resources/my_sections.tex
\usepackage{needspace}



%% file: fig/fig_mse.tex
\begin{tikzpicture}
	\pgfplotsset{grid style={dashed,gray!50}}
	\begin{axis}[
		tick scale binop=\times,
		clip mode=individual,
		minor tick num=1,
		xlabel={Centrality ($\epsilon$)},
		ylabel={Mean Squared Error (MSE)},
		ylabel near ticks,
		width=\columnwidth, 
		height=0.6\columnwidth, 
		xmin=0,xmax=6,ymin=0.6,ymax=1.6,
		grid=both,
		legend style={
			at={(0.98,0.98)}, 
			anchor=north east, 
			legend cell align=left
		}				
	]
	
	\addplot [color=black,only marks,mark=*, mark size=0.5pt]
		table[col sep=comma, x index=0,y expr=\thisrow{mean}]{fig/data/simulated.dat};
	\addlegendentry{Monte Carlo};

	\addplot [color=red,solid,thick,mark=none]
		table[col sep=comma, x index=0,y expr=\thisrow{mean}]{fig/data/computed.dat};
	\addlegendentry{Closed Form};

\end{axis}
\end{tikzpicture}

%% file: fig/fig_dataset.tex
\begin{figure}[t]
\captionsetup[subfigure]{labelformat=empty}
\centering
\begin{subfigure}[b]{.04\columnwidth}
\rotatebox{90}{\quad \small{Art}}
\end{subfigure}
\begin{subfigure}[b]{.15\columnwidth}
\includegraphics[width=1.4cm,height=1.4cm,keepaspectratio]{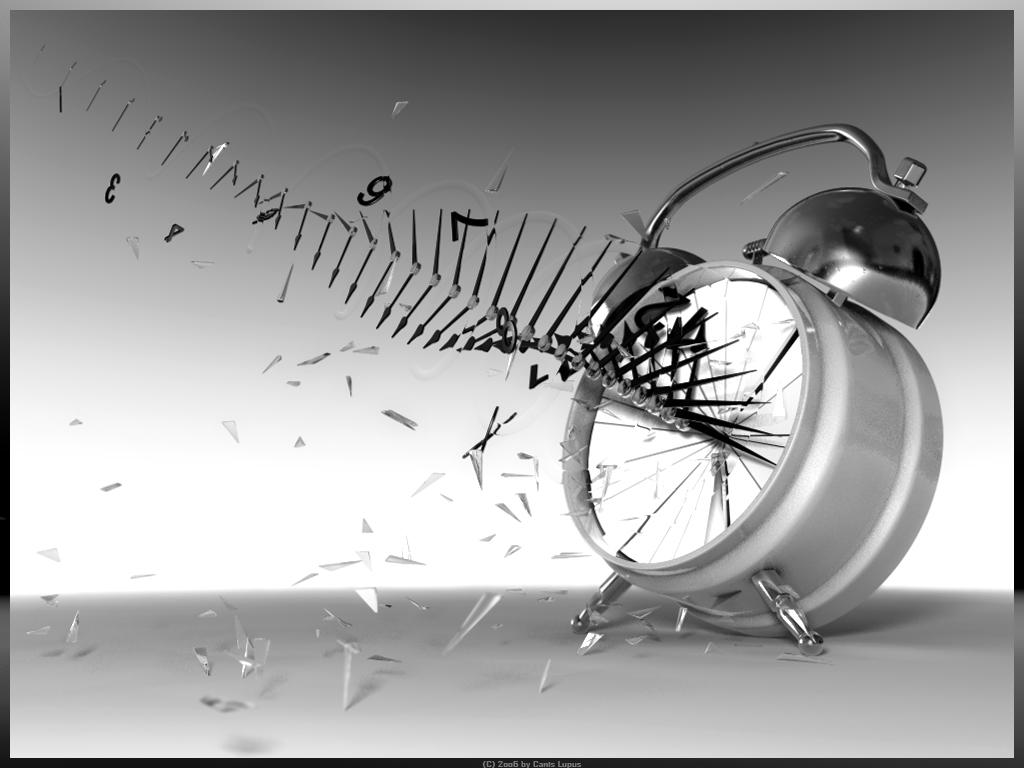}   
\end{subfigure}
\begin{subfigure}[b]{.15\columnwidth}
\includegraphics[width=1.4cm,height=1.4cm,keepaspectratio]{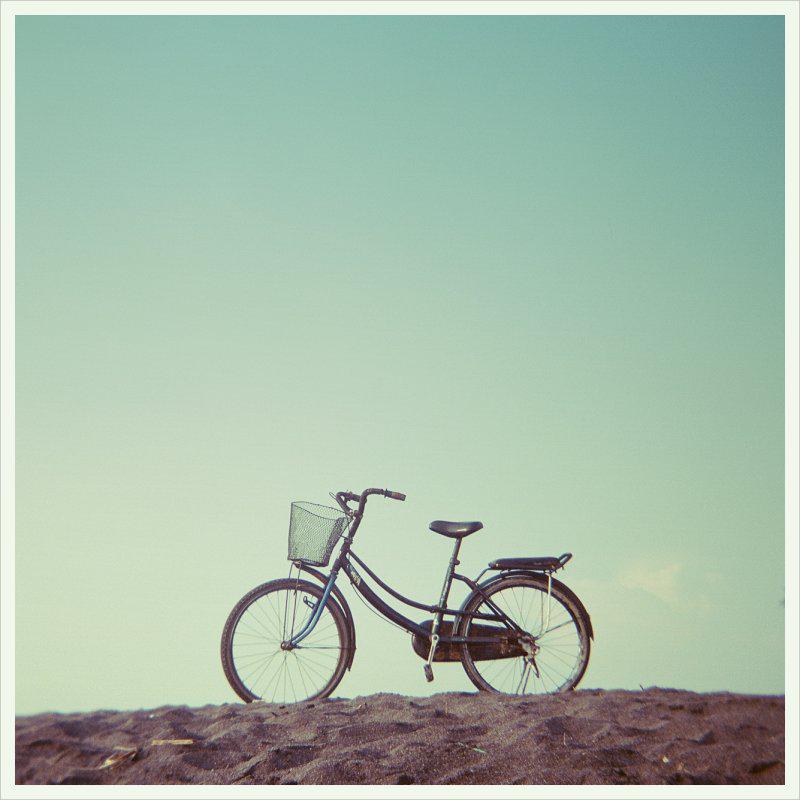}      
\end{subfigure}
\begin{subfigure}[b]{.15\columnwidth}
\includegraphics[width=1.4cm,height=1.4cm,keepaspectratio]{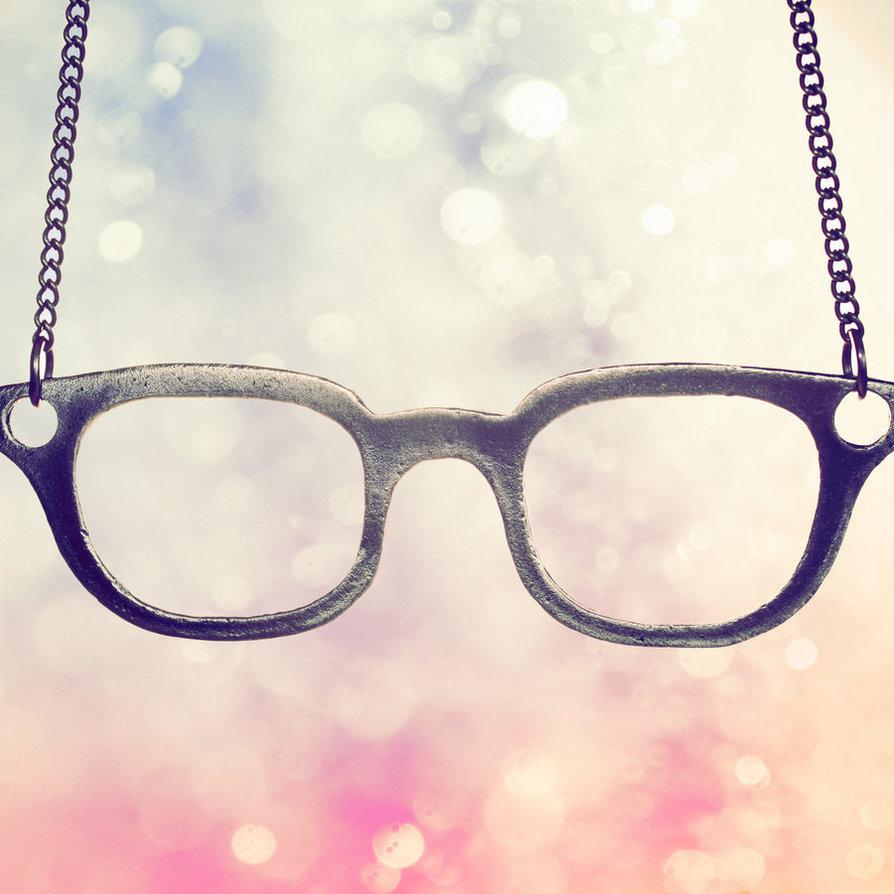}    
\end{subfigure}
\begin{subfigure}[b]{.15\columnwidth}
\includegraphics[width=1.4cm,height=1.4cm,keepaspectratio]{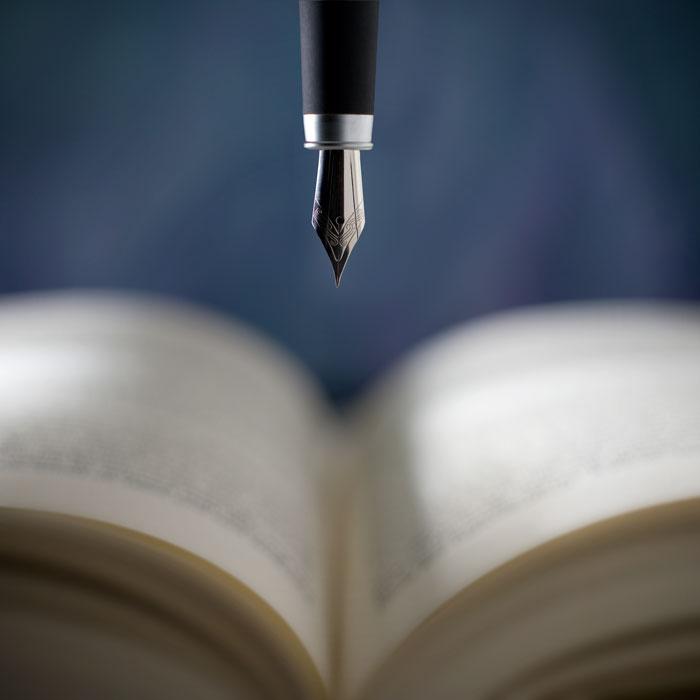}      
\end{subfigure}
\begin{subfigure}[b]{.15\columnwidth}
\includegraphics[width=1.4cm,height=1.4cm,keepaspectratio]{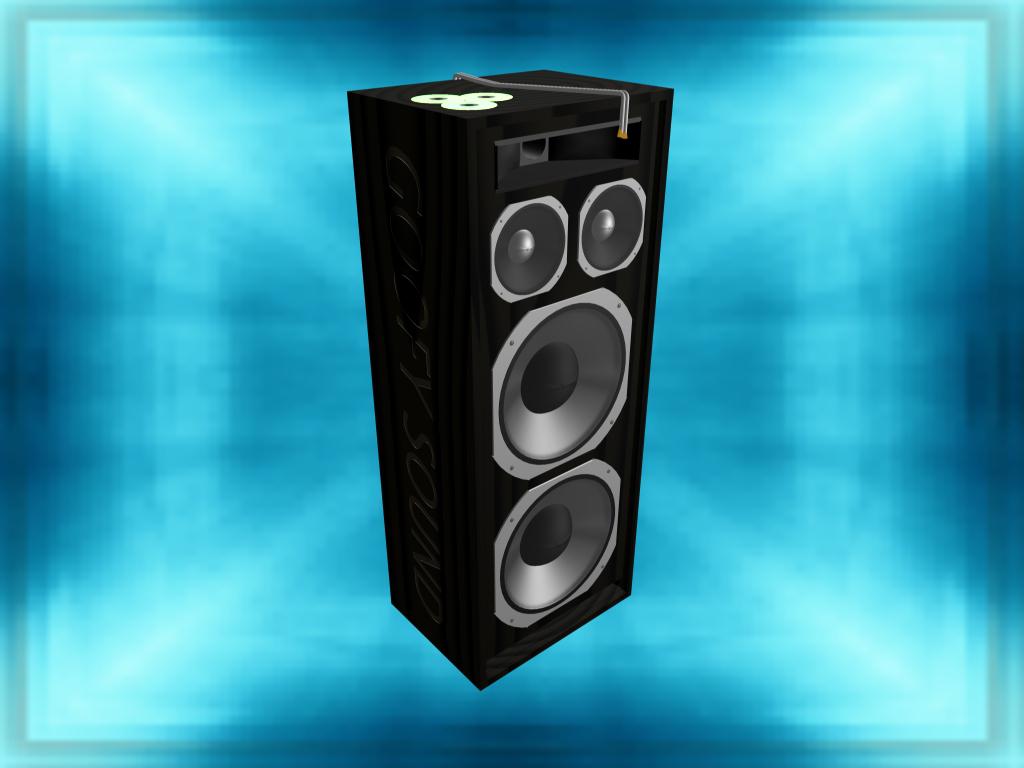}      
\end{subfigure}
\\
\begin{subfigure}[b]{.04\columnwidth}
\rotatebox{90}{\enskip \small{Clipart}}
\end{subfigure}
\begin{subfigure}[b]{.15\columnwidth}
\includegraphics[width=1.4cm,height=1.4cm,keepaspectratio]{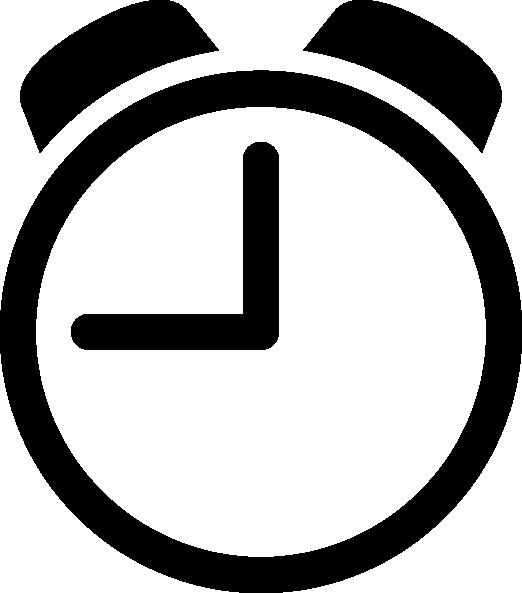}   
\end{subfigure}
\begin{subfigure}[b]{.15\columnwidth}
\includegraphics[width=1.4cm,height=1.4cm,keepaspectratio]{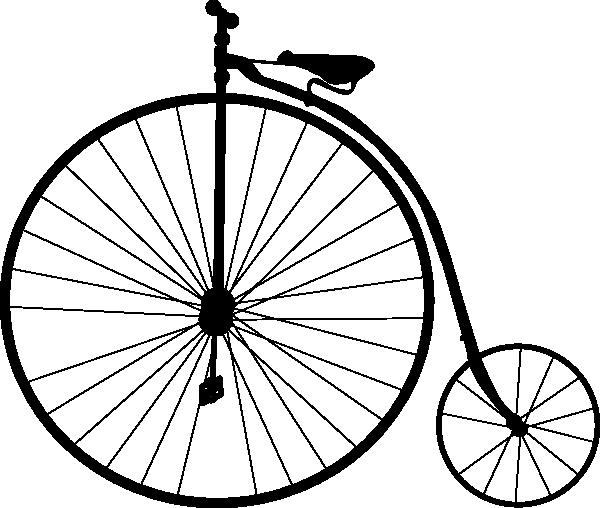}      
\end{subfigure}
\begin{subfigure}[b]{.15\columnwidth}
\includegraphics[width=1.4cm,height=1.4cm,keepaspectratio]{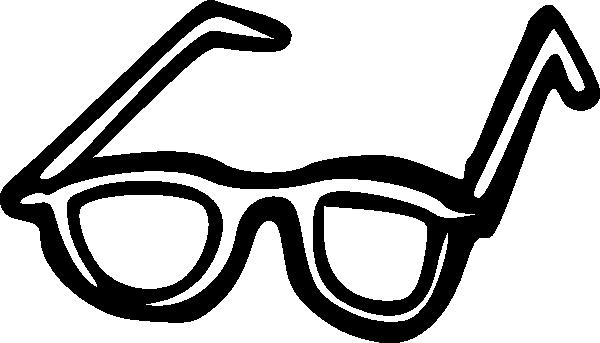}    
\end{subfigure}
\begin{subfigure}[b]{.15\columnwidth}
\includegraphics[width=1.4cm,height=1.4cm,keepaspectratio]{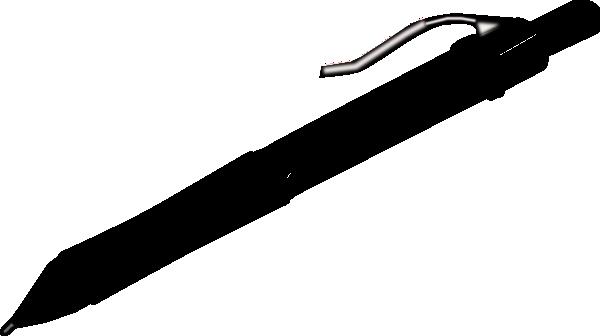}      
\end{subfigure}
\begin{subfigure}[b]{.15\columnwidth}
\includegraphics[width=1.4cm,height=1.4cm,keepaspectratio]{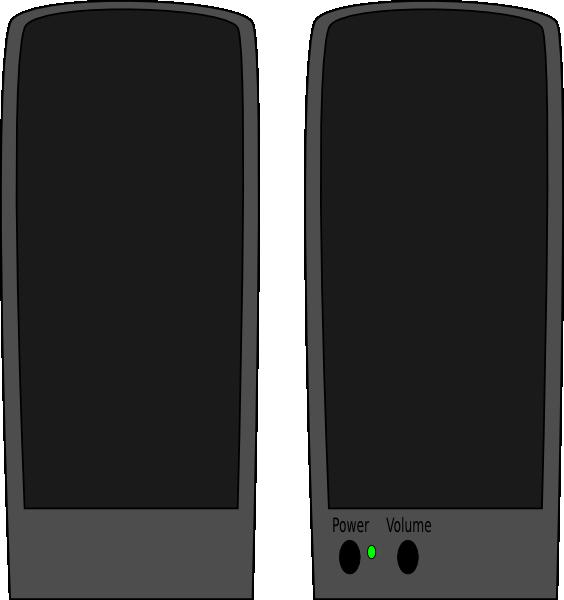}      
\end{subfigure}
\\
\begin{subfigure}[b]{.04\columnwidth}
\rotatebox{90}{\enskip \small{Product}}
\end{subfigure}
\begin{subfigure}[b]{.15\columnwidth}
\includegraphics[width=1.4cm,height=1.4cm,keepaspectratio]{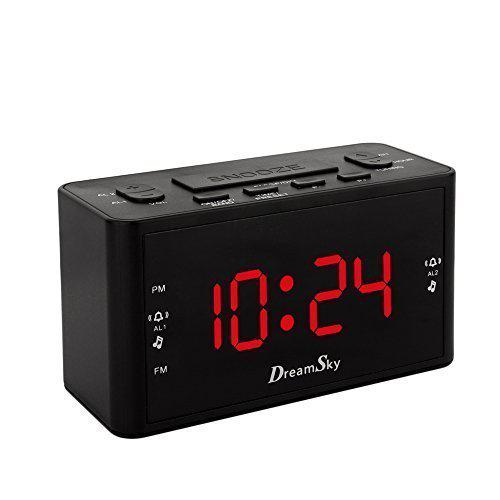}   
\end{subfigure}
\begin{subfigure}[b]{.15\columnwidth}
\includegraphics[width=1.4cm,height=1.4cm,keepaspectratio]{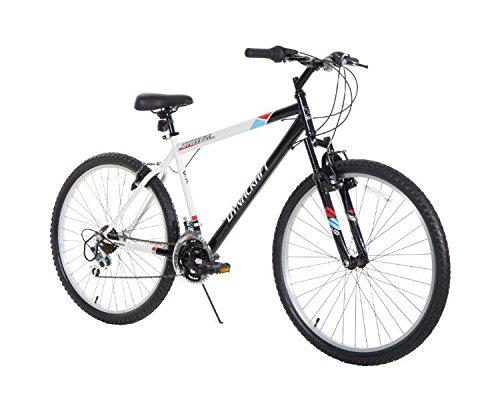}      
\end{subfigure}
\begin{subfigure}[b]{.15\columnwidth}
\includegraphics[width=1.4cm,height=1.4cm,keepaspectratio]{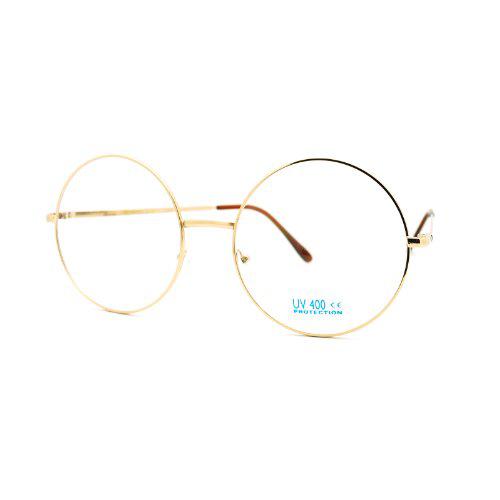}    
\end{subfigure}
\begin{subfigure}[b]{.15\columnwidth}
\includegraphics[width=1.4cm,height=1.4cm,keepaspectratio]{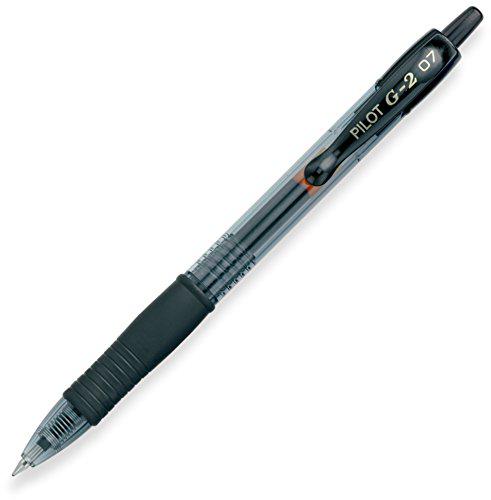}      
\end{subfigure}
\begin{subfigure}[b]{.15\columnwidth}
\includegraphics[width=1.4cm,height=1.4cm,keepaspectratio]{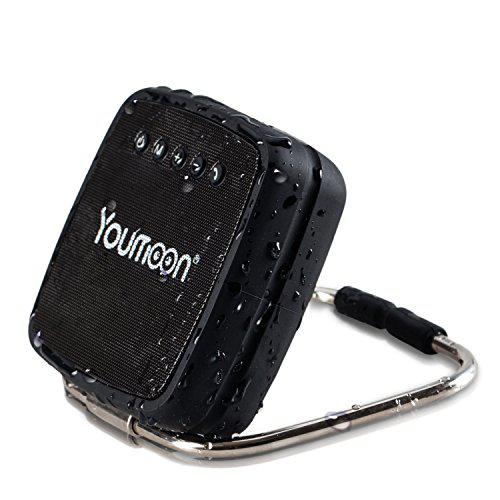}      
\end{subfigure}
\\
\begin{subfigure}[b]{.04\columnwidth}
\rotatebox{90}{\quad\quad \small{World}}
\end{subfigure}
\begin{subfigure}[b]{.15\columnwidth}
\includegraphics[width=1.4cm,height=1.4cm,keepaspectratio]{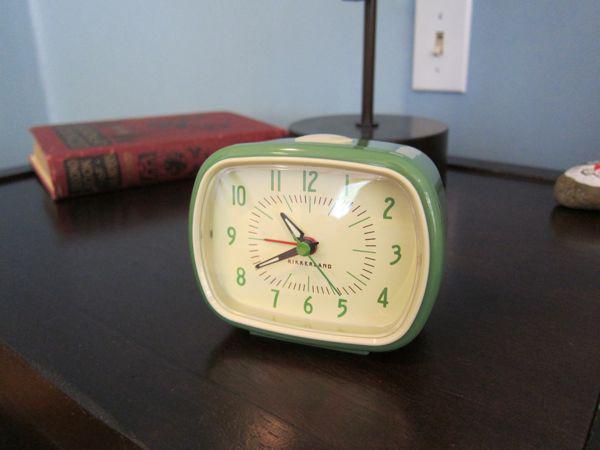}   
\caption{Alarm}
\end{subfigure}
\begin{subfigure}[b]{.15\columnwidth}
\includegraphics[width=1.4cm,height=1.4cm,keepaspectratio]{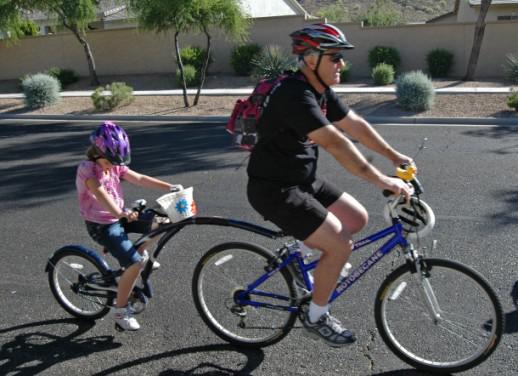}      
\caption{Bike}
\end{subfigure}
\begin{subfigure}[b]{.15\columnwidth}
\includegraphics[width=1.4cm,height=1.4cm,keepaspectratio]{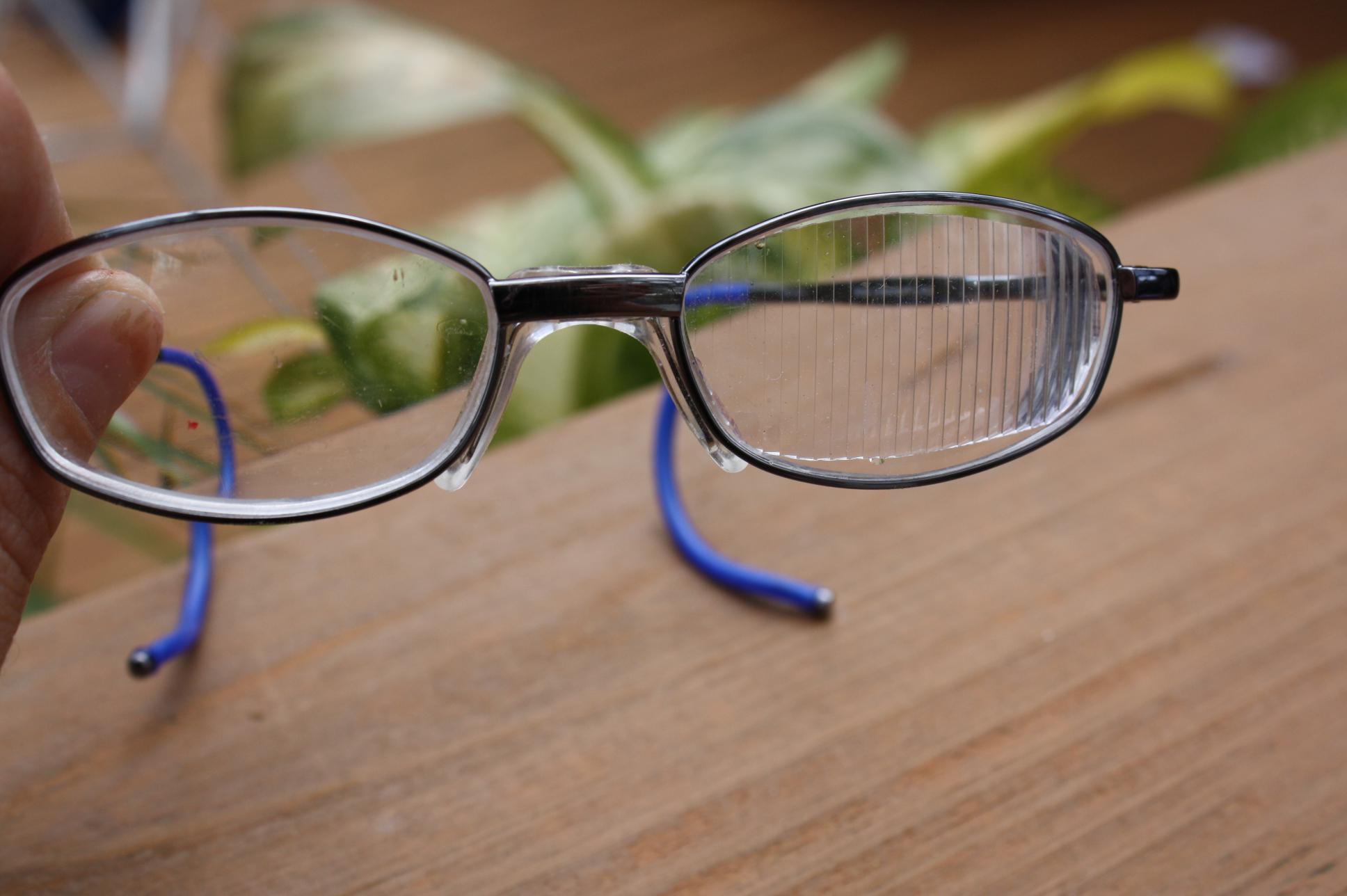}    
\caption{Glasses}
\end{subfigure}
\begin{subfigure}[b]{.15\columnwidth}
\includegraphics[width=1.4cm,height=1.4cm,keepaspectratio]{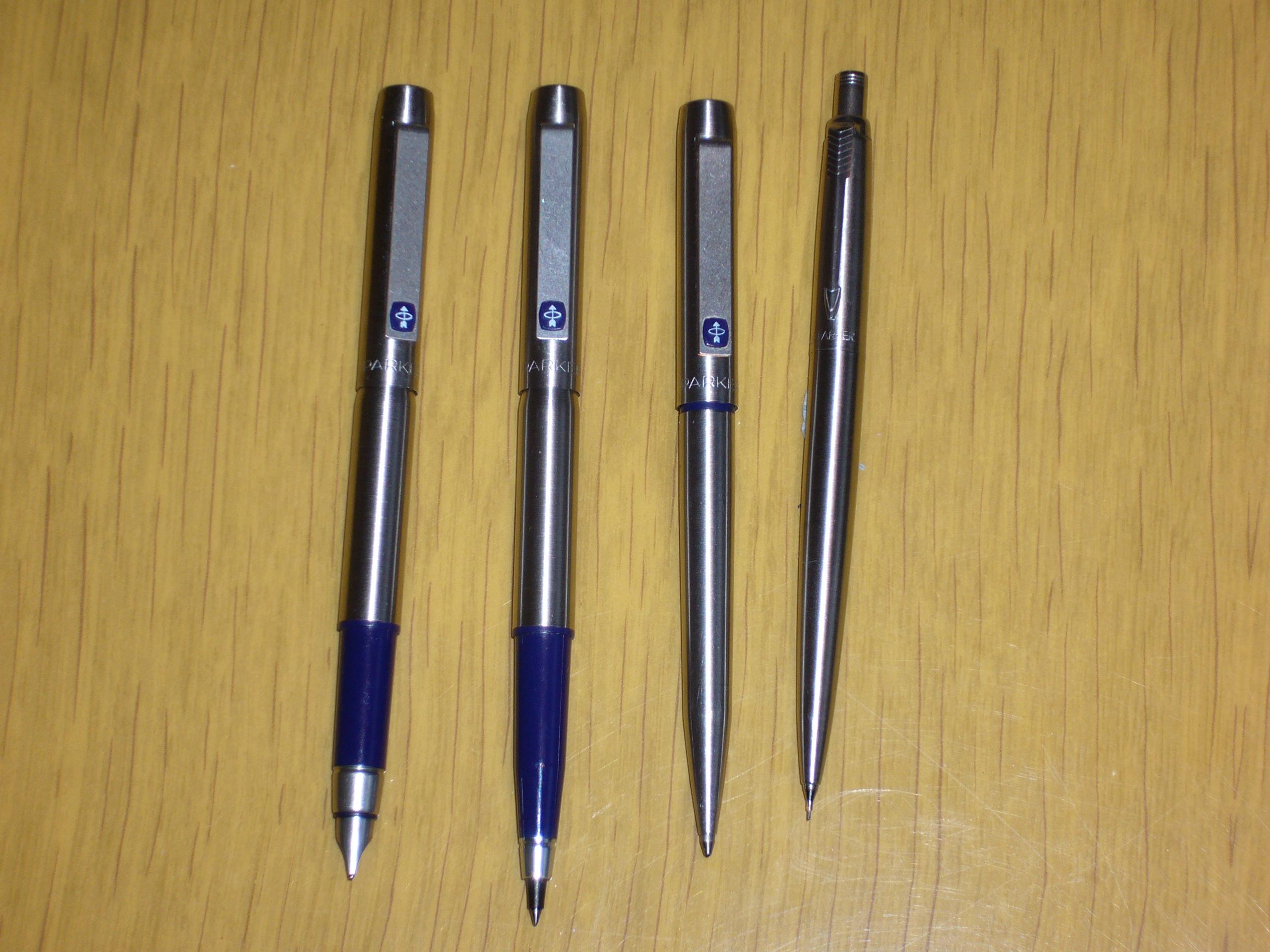}      
\caption{Pen}
\end{subfigure}
\begin{subfigure}[b]{.15\columnwidth}
\includegraphics[width=1.4cm,height=1.4cm,keepaspectratio]{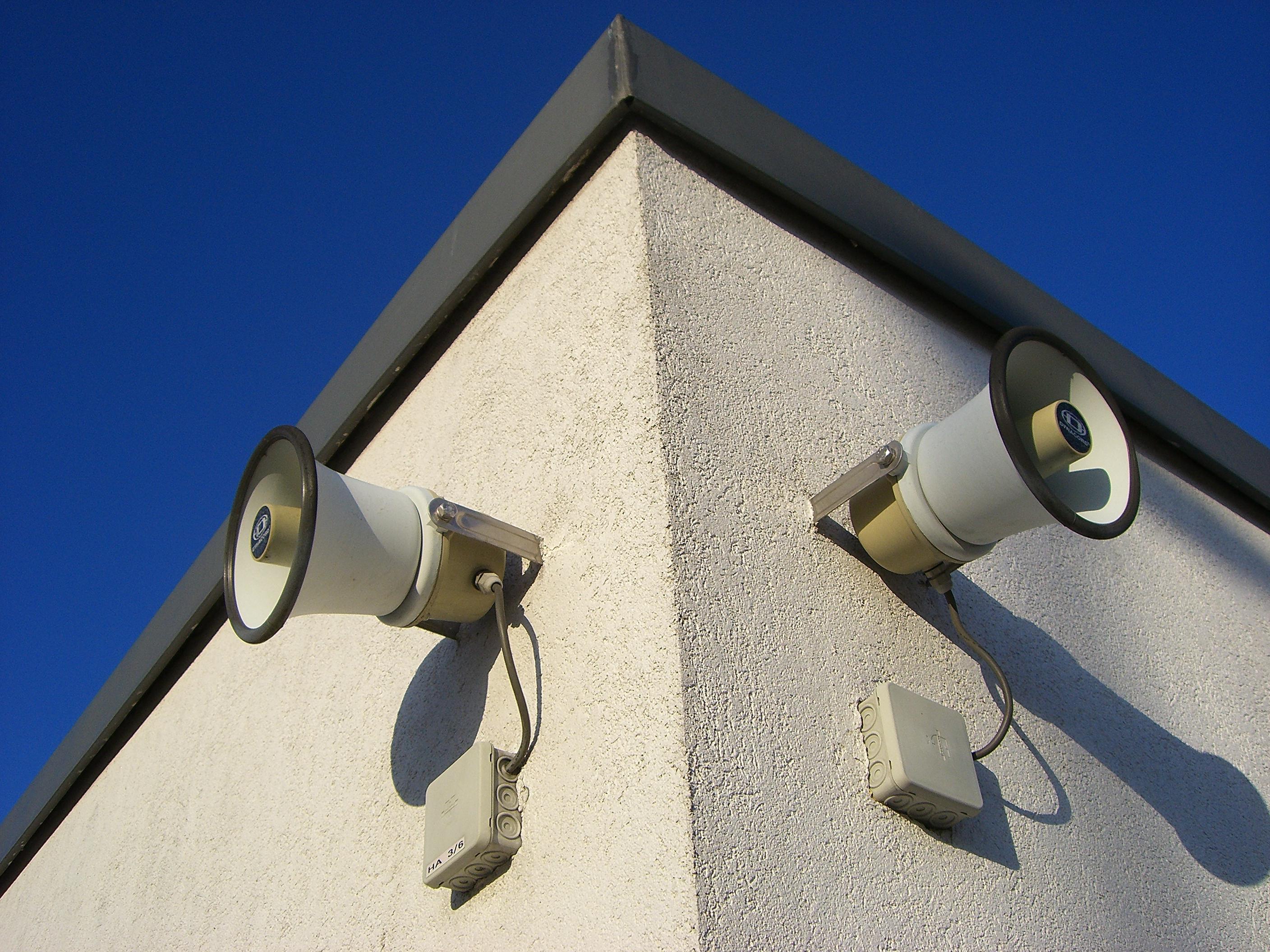}
\caption{Speaker}
\end{subfigure}
\caption{Example images from $5$ of the $65$ categories from the $4$ domains composing the Office-Home dataset \cite{venkateswara2017Deep}. The $4$ domains are Art, Clipart, Product and Real World. In total, the dataset contains $15{,}500$ images of different sizes and shapes.}
\label{fig:Dataset}
\end{figure}

%% file: fig/fig_epsilons_mean.tex
\begin{tikzpicture}
	\pgfplotsset{grid style={dashed,gray!50}}

	\begin{semilogxaxis}[
		tick scale binop=\times,
        label style={font=\small},			        		
		minor tick num=1,
		xlabel={Centrality ($\epsilon$)},
		ylabel={Accuracy ($\%$)},
		ylabel near ticks,
		width=\columnwidth, 
        xtick={0.00000001, 0.0000001, 0.000001, 0.00001, 0.0001, 0.001, 0.01, 0.1},
        xticklabels={$0$,$10^{-7}$,$10^{-6}$,$10^{-5}$,$10^{-4}$,$10^{-3}$,$10^{-2}$,$10^{-1}$},
		height=0.5\columnwidth, 
		xmin=0.00000001,xmax=0.1,ymin=32,ymax=38,
		grid=both,
		axis line style={draw=none}
	]
	
	
\addplot [name path=upperStd,draw=none] 
				table[col sep=comma, x index=0,y expr=\thisrow{Mean}+ \thisrow{Std}] {fig/data/Mean_Complete.dat};
\addplot [name path=lowerStd,draw=none] 
				table[col sep=comma, x index=0,y expr=\thisrow{Mean}-\thisrow{Std}] {fig/data/Mean_Complete.dat};
\addplot [fill=red,opacity=0.2] fill between[of=upperStd and lowerStd];

\addplot [color=red,solid,thick,mark=none,mark options={red,solid}]
				table[col sep=comma, x index=0,y expr=\thisrow{Mean}]{fig/data/Mean_Line.dat};

\addplot [color=red, dotted,thick] 
		table[col sep=comma, x index=0,y expr=\thisrow{Mean}] {fig/data/Mean_Line_Consensus.dat};

\addplot [color=red, dashed,thick] 
	table[col sep=comma, x index=0,y expr=\thisrow{Mean}] {fig/data/Mean_Line_Agnostic.dat};	

\addplot [color=red, only marks, mark=*,mark options={fill=red,scale=0.5, color =red}] 
	table[col sep=comma, x index=0,y expr=\thisrow{Mean}] {fig/data/Mean_Complete_Dots.dat};

\end{semilogxaxis}

\begin{axis}[
		axis x discontinuity=crunch,
		xlabel={},
		ylabel={},
		ylabel near ticks,
		width=\columnwidth, 
        xticklabels={,,},
        yticklabels={,,},        
        ytick style={draw=none},
        xtick style={draw=none},
		height=0.5\columnwidth, 
		xmin=0.00000001,xmax=0.1,ymin=32,ymax=38,
		grid=none
	]

\addplot [color=red, only marks, mark=*,mark options={fill=red,scale=0.5, color =red}] 
	table[col sep=comma, x index=0,y expr=\thisrow{Mean}] {fig/data/Mean_Complete_Dots_Axis.dat};

\end{axis}	

\end{tikzpicture}

%% file: fig/fig_bars_agnostic.tex
\begin{tikzpicture}

	\pgfplotsset{grid style={dashed,gray!50}}
	\begin{axis}[
		xbar,
		bar shift=0pt,
		bar width=7pt,		
		area legend,
		minor tick num=0,
		ytick={1,2,3,4},
		yticklabels={,,},
		xlabel={Improvement ($\%$)},
		ylabel near ticks,
		width=1.1\columnwidth, 
		height=0.4\columnwidth, 
		ymin=0.25,ymax=4.75, xmin=0,xmax=250, 
		grid=both,
		legend style={
			at={(0.5,1.06)}, 
			anchor=south, 
			legend cell align=left,
			legend columns=4
		}		
	]

	\addplot [draw=red, fill=red!30, nodes near coords={$186.61\%$}] 
		table[col sep=comma, y expr=\thisrow{IndexArt}, x expr=\thisrow{Art}] {fig/data/Agnostic_bar.dat};
	\addlegendentry{Art};

	\addplot [draw=blue, fill=blue!30, nodes near coords={$39.20\%$}] 
		table[col sep=comma,  y expr=\thisrow{IndexClipart}, x expr=\thisrow{Clipart}] {fig/data/Agnostic_bar.dat};
	\addlegendentry{Clipart};

	\addplot [draw=green, fill=green!30, nodes near coords={$26.98\%$}] 
		table[col sep=comma,  y expr=\thisrow{IndexProduct}, x expr=\thisrow{Product}] {fig/data/Agnostic_bar.dat};
	\addlegendentry{Product};

	\addplot [draw=black, fill=black!30, nodes near coords={$54.82\%$}] 
		table[col sep=comma,  y expr=\thisrow{IndexReal}, x expr=\thisrow{RealWorld}] {fig/data/Agnostic_bar.dat};
	\addlegendentry{Real World};

\end{axis}

\end{tikzpicture}

%% file: fig/fig_bars_consensus.tex
\begin{tikzpicture}

	\pgfplotsset{grid style={dashed,gray!50}}
	\begin{axis}[
		xbar,
		bar shift=0pt,
		bar width=7pt,		
		area legend,
		minor tick num=0,
		ytick={1,2,3,4},
		yticklabels={,,},
		xlabel={Improvement ($\%$)},
		ylabel near ticks,
		width=1.1\columnwidth, 
		height=0.4\columnwidth, 
		ymin=0.25,ymax=4.75, xmin=0,xmax=50, 
		grid=both,
		legend style={
			at={(0.5,1.06)}, 
			anchor=south, 
			legend cell align=left,
			legend columns=4
		}		
	]

	\addplot [draw=red, fill=red!30, nodes near coords={$26.38\%$}] 
		table[col sep=comma, y expr=\thisrow{IndexArt}, x expr=\thisrow{Art}] {fig/data/Consensus_bar.dat};
	
	\addplot [draw=blue, fill=blue!30, nodes near coords={$5.13\%$}] 
		table[col sep=comma,  y expr=\thisrow{IndexClipart}, x expr=\thisrow{Clipart}] {fig/data/Consensus_bar.dat};

	\addplot [draw=green, fill=green!30, nodes near coords={$6.65\%$}] 
		table[col sep=comma,  y expr=\thisrow{IndexProduct}, x expr=\thisrow{Product}] {fig/data/Consensus_bar.dat};

	\addplot [draw=black, fill=black!30, nodes near coords={$0.74\%$}] 
		table[col sep=comma,  y expr=\thisrow{IndexReal}, x expr=\thisrow{RealWorld}] {fig/data/Consensus_bar.dat};

\end{axis}

\end{tikzpicture}